\documentclass[11pt]{article}

\usepackage[utf8]{inputenc}
\usepackage[T1]{fontenc}
\usepackage{amsmath,amssymb}
\usepackage{graphicx}
\usepackage{booktabs}
\usepackage{hyperref}
\usepackage{algorithm}
\usepackage{algorithmic}
\usepackage{geometry}
\usepackage{xcolor}
\usepackage{enumitem}
\usepackage{authblk}
\usepackage{tikz}
\usetikzlibrary{positioning,arrows.meta,shapes.geometric,fit,backgrounds,calc}

\hypersetup{
    colorlinks=true,
    linkcolor=blue!70!black,
    citecolor=blue!70!black,
    urlcolor=blue!70!black,
}

\title{CellDX AI Autopilot: Agent-Guided Training and Deployment\\of Pathology Classifiers}

\author[1]{Alexey Pchelnikov\thanks{\texttt{alex@hist.ai}}}
\author[1]{Aleksei Pchelnikov\thanks{\texttt{apchelnikov@hist.ai}}}
\affil[1]{HistAI}

\date{\today}

\begin{document}
\maketitle

\begin{abstract}
Training AI models for computational pathology currently requires access to expensive whole-slide-image datasets, GPU infrastructure, deep expertise in machine learning, and substantial engineering effort. We present CellDX AI Autopilot, a platform that lets users---from pathologists with no ML background to ML practitioners running many parallel experiments---train, evaluate, and deploy whole-slide image classifiers through natural language interaction with an AI agent. The platform provides a structured set of agent skills that guide the user through dataset curation, automated hyperparameter tuning, multi-strategy model comparison, and human-in-the-loop deployment---all on a pre-built dataset of over 32{,}000 cases and 66{,}000 H\&E-stained whole-slide images with pre-extracted features. We describe the agent skill architecture, the underlying Multiple Instance Learning (MIL) training framework supporting four classification strategies, and an iterative pairwise hyperparameter search (grid or seeded random) that reduces tuning cost by over 30$\times$ compared to exhaustive search. CellDX AI Autopilot is, to our knowledge, the first system to expose pathology-specialized agent skills and a pathology-specialized training platform to general-purpose AI agents (e.g.\ any LLM-based agent runtime), delivering end-to-end automated model training without requiring the agent itself to be domain-specific. The platform addresses both the ML-expertise bottleneck that limits adoption in diagnostic pathology and the engineering bottleneck that limits how many experiments a researcher can run cost-effectively.
\end{abstract}

\section{Introduction}

Artificial intelligence has demonstrated remarkable potential in computational pathology, with models achieving expert-level performance on tasks such as cancer detection~\cite{campanella2019clinical}, subtype classification~\cite{lu2021data}, and biomarker prediction. However, the translation of these results into routine clinical practice remains slow. A primary bottleneck is not the availability of methods, but the expertise required to apply them: training a slide-level classifier requires knowledge of feature extraction, MIL aggregation strategies, hyperparameter optimization, class imbalance handling, and model validation---skills that few pathologists possess.

Existing tools fall into two categories. Commercial platforms (PathAI, Paige.AI) develop models internally and offer them as black-box products; pathologists consume predictions but cannot build custom classifiers for their own research questions. Open-source frameworks (CLAM~\cite{lu2021data}, SlideFlow~\cite{dolezal2024slideflow}, PathML) provide the building blocks but require significant programming and ML expertise to use effectively. Neither category empowers pathologists to independently train models for novel classification tasks.

The emergence of large language model (LLM) agents capable of executing multi-step workflows~\cite{wu2023autogen,huang2024mlagentbench} opens a new possibility: pair a general-purpose AI agent with a domain-specialized skill set and training platform, and let the agent guide the user through the entire process via natural language conversation. The agent itself does not need to know pathology---the pathology knowledge lives in the skills and the platform---so the same general-purpose agent runtime can be retargeted to other scientific domains by swapping the skills. Such an agent is useful both to clinicians without ML training---who can describe a classification objective in plain language and obtain a deployable model---and to experienced ML practitioners, who can use it to parallelize many configurations or sweep hyperparameters across strategies without writing scaffolding code. Recent work on LLM agents for data science~\cite{hong2024data,guo2024dsagent} has shown that agents can automate ML pipelines on standard benchmarks, but no published system applies this approach to the specialized domain of computational pathology.

We present \textbf{CellDX AI Autopilot}, a platform that bridges this gap. The system provides:

\begin{enumerate}[leftmargin=*]
    \item \textbf{AI agent skills for pathology ML.} A structured set of skills that enable an AI agent to navigate the full training pipeline---from dataset validation to model deployment---with domain-appropriate guardrails and sensible defaults.
    \item \textbf{Automated training infrastructure.} Cloud-based GPU training with four MIL strategies, automated hyperparameter tuning, real-time metric monitoring, and reproducible evaluation.
    \item \textbf{Human-in-the-loop deployment.} A deliberate separation between training and deployment, requiring explicit user approval before any model goes live---reflecting the safety requirements of clinical AI.
\end{enumerate}

The platform currently operates on H\&E-stained whole-slide images for classification tasks, with pre-extracted features from a dataset of over 32{,}000 cases and 66{,}000 slides. We describe the architecture, the agent skill design, the training framework, and discuss the current limitations and future directions.

\section{Related Work}

\paragraph{AutoML in medical imaging.} Automated machine learning has seen broad adoption in general computer vision, with systems like Auto-sklearn~\cite{feurer2022autosklearn}, AutoKeras, and Google Vertex AI automating model selection and hyperparameter tuning. In medical imaging, MONAI Auto3DSeg~\cite{myronenko2023monai} automates architecture search for 3D segmentation tasks in radiology. However, none of these systems support the MIL paradigm required for whole-slide image classification, where each sample is a variable-length bag of thousands of patch features rather than a single fixed-size image.

\paragraph{MIL frameworks for pathology.} Several frameworks provide MIL implementations for computational pathology. CLAM~\cite{lu2021data} introduced clustering-constrained attention MIL with instance-level supervision. SlideFlow~\cite{dolezal2024slideflow} offers an end-to-end pipeline from slide processing to MIL training. STAMP provides a structured MIL framework. While these tools are valuable for ML researchers, they require command-line proficiency, understanding of MIL concepts, and manual hyperparameter selection---skills that are uncommon among practicing pathologists.

\paragraph{Pathology foundation models.} Recent foundation models for pathology---including UNI~\cite{chen2024uni}, CONCH~\cite{lu2024conch}, Virchow~\cite{vorontsov2024virchow}, Prov-GigaPath~\cite{xu2024gigapath}, Hibou~\cite{nechaev2024hibou}, and GenBio-PathFM~\cite{genbio2026pathfm}---provide powerful feature extractors that dramatically reduce the complexity of downstream classification. By producing high-quality patch-level representations, these models shift the bottleneck from feature learning to aggregation and classification---exactly the task that CellDX AI Autopilot automates.

\paragraph{LLM agents for ML automation.} The use of LLM-based agents for automated machine learning is a rapidly emerging field. MLAgentBench~\cite{huang2024mlagentbench} evaluates LLM agents on ML experimentation tasks. DS-Agent~\cite{guo2024dsagent} uses case-based reasoning to automate data science workflows. AIDE demonstrates autonomous ML engineering on competitive benchmarks. AutoGen~\cite{wu2023autogen} provides a multi-agent framework for complex task automation. However, none of these systems incorporate domain-specific knowledge about pathology, MIL, or the clinical requirements of medical AI validation.

\paragraph{The gap.} To our knowledge, no published system combines (1) pathology-specialized agent skills consumable by an off-the-shelf general-purpose LLM agent, (2) a pathology-specialized training platform (sharded feature store, MIL training framework, autoscaled GPU compute), and (3) human-in-the-loop deployment controls. CellDX AI Autopilot addresses this gap: the agent runtime stays general-purpose, while the domain knowledge lives entirely in the skills and the platform. A further practical barrier is data acquisition itself: commercial whole-slide-image datasets sufficient to train a clinically useful classifier routinely cost tens to hundreds of thousands of USD per cohort, putting them out of reach of most research groups and a hard blocker for one-off feasibility studies. CellDX AI Autopilot ships with embedded, no-cost access to an ever-growing curated dataset of pre-extracted features (currently 32{,}000+ cases / 66{,}000+ slides; see Section~\ref{sec:training}), so users pay only for the GPU compute their own jobs consume.

\section{Agent Skill Architecture}
\label{sec:skills}

The core design principle of CellDX AI Autopilot is that the complexity of ML model training should be managed by the AI agent, not by the user. To achieve this, the platform exposes its capabilities through a set of structured \textit{skills}---machine-readable documents that tell the agent what actions are available, what parameters are required, and what guardrails apply. The skills are open and version-controlled at \url{https://github.com/histai/skillsets}, so that the platform's training workflow can be inspected, audited, and adapted by the community.

\subsection{Skill Design}

Each skill is a self-contained specification that describes:

\begin{itemize}[leftmargin=*]
    \item \textbf{Available actions} --- the API endpoints the agent can call, with their parameters, defaults, and valid ranges.
    \item \textbf{Recommended workflows} --- step-by-step procedures for common tasks (e.g., the full hyperparameter tuning $\rightarrow$ strategy comparison $\rightarrow$ deployment pipeline).
    \item \textbf{Domain constraints} --- rules that encode pathology ML best practices. For example: always validate that slide features exist before building a cohort; never deploy a model without explicit user approval; use clinical language (not ML jargon) in user-facing model descriptions.
    \item \textbf{Error handling guidance} --- how to interpret and respond to error conditions (insufficient balance, failed jobs, convergence issues).
\end{itemize}

The platform currently provides two complementary skills:

\paragraph{AI Model Trainer skill.} Guides the agent through the full training lifecycle: discovering available configuration options, gathering user requirements, validating the cohort against available features, creating training sessions, submitting jobs (single training, HP tuning, or strategy comparison), and deploying approved models.

\paragraph{Training Monitor skill.} Enables the agent to track running jobs, interpret training metrics (identifying overfitting, plateau, or convergence), present results in accessible formats, and recommend adjustments (stopping early, reducing learning rate, adding regularization).

\subsection{Agent--User Interaction Pattern}

A typical interaction proceeds as follows:

\begin{enumerate}[leftmargin=*]
    \item The pathologist describes their classification objective in natural language (e.g., ``I want to classify breast biopsies as lobular carcinoma versus invasive ductal carcinoma'').
    \item The agent uses the Trainer skill to validate the available slide data, checking which slides have pre-extracted features.
    \item The agent proposes a training plan: cohort split, recommended strategy, hyperparameter ranges. The user can accept, modify, or ask questions.
    \item Training runs on cloud GPU infrastructure. The agent uses the Monitor skill to track progress and report results.
    \item When training completes, the agent presents the results and asks the user whether to deploy. No model is deployed without explicit approval.
\end{enumerate}

\begin{figure}[ht]
\centering
\begin{tikzpicture}[
    every node/.style={font=\small},
    box/.style={
        draw, rounded corners=2pt,
        minimum width=46mm, minimum height=12mm,
        align=center,
        inner sep=4pt,
    },
    user/.style    ={box, fill=blue!8,    draw=blue!50!black},
    agent/.style   ={box, fill=orange!12, draw=orange!60!black},
    api/.style     ={box, fill=green!10,  draw=green!50!black},
    cloud/.style   ={box, fill=purple!8,  draw=purple!60!black},
    artifact/.style={box, fill=gray!10,   draw=gray!60!black, minimum width=42mm},
    flow/.style   ={-{Latex[length=2.2mm]}, thick},
    revflow/.style={-{Latex[length=2.2mm]}, thick, dashed},
    lbl/.style    ={font=\scriptsize, fill=white, inner sep=1.5pt, midway},
    node distance=18mm and 22mm,
]

\node[user]                     (user)   {Pathologist /\\ML practitioner};
\node[agent,  below=of user]    (agent)  {AI Agent\\(skills loaded)};
\node[api,    below=of agent]   (api)    {CellDX API\\jobs $\cdot$ sessions $\cdot$ runs};
\node[cloud,  below=of api]     (gpu)    {Azure ML GPU compute\\(autoscaled cluster)};

\node[artifact, left=of agent]  (skills)  {Agent Skills\\\scriptsize\texttt{github.com/histai/skillsets}};
\node[artifact, left=of gpu]    (metrics) {Metric history\\(MongoDB)};
\node[artifact, right=of api]   (widget)  {Deployed widget};

\draw[flow] (user)  -- node[lbl, right]{1.~goal in plain language} (agent);
\draw[flow] (agent) -- node[lbl, left] {2.~consult}                (skills);
\draw[flow] (agent) -- node[lbl, right]{3.~submit jobs}            (api);
\draw[flow] (api)   -- node[lbl, right]{4.~train}                  (gpu);
\draw[flow] (gpu)   -- node[lbl, above]{5.~per-epoch metrics}      (metrics);

\draw[flow] (metrics.east) -- node[lbl, above, pos=0.55]{} (api.west);

\draw[revflow] (api.east) to[bend right=55] node[lbl, right=1mm]{6.~results} (agent.east);
\draw[revflow] (agent.east) to[bend right=55] node[lbl, right=1mm]{7.~summary \& ask} (user.east);

\draw[flow] (user.west) to[bend right=55] node[lbl, left=1mm]{8.~approve} (agent.west);
\draw[flow] (api.east) -- node[lbl, above]{9.~deploy} (widget.west);

\end{tikzpicture}
\caption{Agent--user interaction pattern. The main spine on the left is the forward path (solid): the user states a goal in natural language, the agent consults its skill specifications, submits jobs through the CellDX API, and those jobs run on the autoscaled Azure ML GPU cluster, which streams per-epoch metrics back into MongoDB and through the API. Dashed arrows (steps 6--7) carry results back: the API surfaces metric summaries to the agent, which translates them into a clinical summary for the user. Deployment requires an explicit user-approval step (8); only then does the agent issue the deploy call (9) and the widget becomes a serving model.}
\label{fig:agent-flow}
\end{figure}
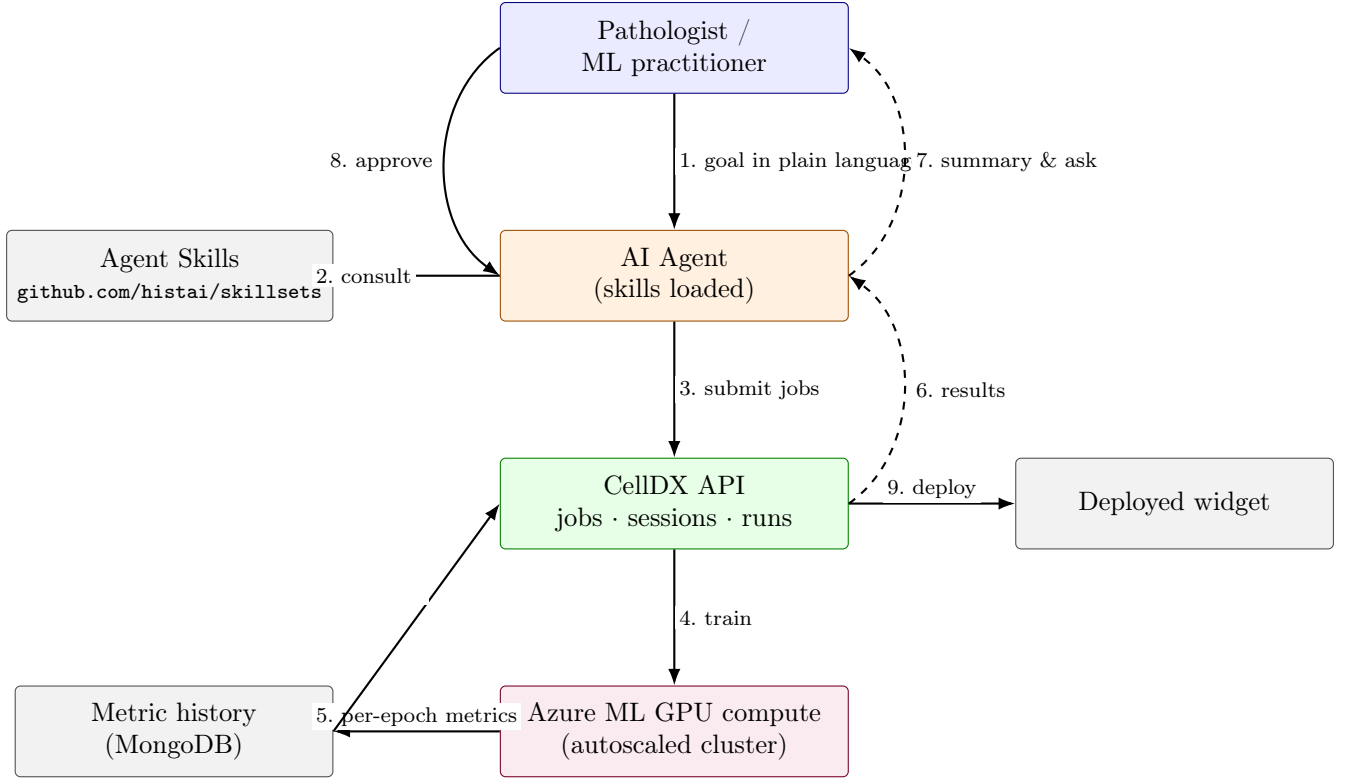

This interaction pattern keeps the pathologist in control of clinical decisions (what to classify, which results are acceptable, whether to deploy) while delegating the technical decisions (which aggregation strategy, what learning rate, how many epochs) to the agent. Figure~\ref{fig:agent-flow} summarises the message flow.

\subsection{Guardrails and Safety}

The skill specifications encode several safety constraints:

\begin{itemize}[leftmargin=*]
    \item \textbf{No automatic deployment.} The agent must present training results and obtain explicit user approval before deploying any model.
    \item \textbf{No internal exposure.} Model architecture details (aggregator type, hidden dimensions, dropout rates) are never included in user-facing descriptions. Widget titles and descriptions use clinical terminology only.
    \item \textbf{Feature validation.} The agent must verify that all slides in the cohort have pre-extracted features before submitting a training job, preventing silent failures.
    \item \textbf{Minimum cohort size.} The agent is instructed not to submit training jobs when the available data is insufficient (fewer than 20 samples per class).
\end{itemize}

\section{Training Framework}
\label{sec:training}

Behind the agent interface, CellDX AI Autopilot provides a MIL training framework that handles the technical complexity of model training.

\subsection{System Architecture}

The platform consists of two independently deployed components that communicate asynchronously via structured log messages:

\begin{itemize}[leftmargin=*]
    \item An \textbf{orchestration API} that manages job submissions, tracks training state, and provides the REST endpoints that the agent skills call. The API persists job state and metrics in a document database and submits training commands to a cloud ML compute service. A background poller runs on a 30-second interval, fetching the trainer's stdout logs from blob storage and parsing structured JSON lines (prefixed with \texttt{[trainer]}) to extract per-epoch metrics. This log-based streaming architecture decouples the trainer from the API: the trainer writes to stdout, the cloud platform captures it to blob storage, and the API reads it asynchronously---no direct network connection is required between them.
    \item A \textbf{GPU trainer} container that executes on cloud GPU compute. Each training job receives a complete JSON configuration (cohort, hyperparameters, strategy, data source) and runs independently. At each epoch, the trainer emits a JSON object containing all training and validation metrics to stdout. Checkpoints are written to a persistent output mount that survives after the GPU instance is released, ensuring trained model weights are available for deployment.
\end{itemize}

The system follows a protocol-based design: data loading and metric tracking are defined as abstract protocols, allowing the trainer to operate identically whether loading from a single H5 file, multiple tissue-specific files, or the sharded feature store. This same extensibility applies to tracking backends and, in the future, to alternative feature extractors.

\paragraph{Deployment environment.} All components are deployed inside a secured Microsoft Azure environment: the orchestration API runs as a managed Container App with secrets stored in a managed vault, training data and feature shards live in private Azure Blob Storage accessed through a managed identity, and the document database is a private Cosmos DB endpoint. No service exposes raw storage credentials to the user-facing path, and every API call is authenticated against the platform's existing identity layer.

\paragraph{GPU compute and autoscaling.} Training jobs run on a cluster of GPU instances managed by the cloud ML service. The cluster scales automatically with load: idle instances are scaled down when no jobs are queued, and additional instances spin up as more jobs are submitted (subject to a configurable cluster ceiling). This elasticity is what makes parallel multi-strategy or HP-tuning sweeps economically viable---users pay for compute only while their jobs are actually running, and there is no idle-capacity overhead between training campaigns.

\paragraph{Compute-only pricing for users.} HistAI provides access to the curated commercial feature store---over 32{,}000 cases and 66{,}000 slides, pre-tiled and pre-encoded---free of charge to all CellDX platform users. Users pay only for the GPU compute they actually consume. This pricing structure is designed to remove the largest single cost of computational-pathology research (acquiring and processing whole-slide images) from the equation, so that the marginal cost of an additional experiment is just the GPU hours it consumes.

\subsection{Feature Representation and Data Pipeline}

The platform operates on pre-extracted patch-level features from Hibou-L~\cite{nechaev2024hibou}, a vision transformer pretrained on 1.2 million histopathology images. Source slides are scanned at $20\times$ magnification, and patches are extracted from the highest-detail level of the pyramidal whole-slide image---i.e., the base $20\times$ level rather than any downsampled tier---so that the encoder operates on cellular-scale detail. Each H\&E-stained slide is tessellated into non-overlapping 224$\times$224 pixel patches, and foreground patches are identified via tissue masking. The feature extractor produces a 1024-dimensional embedding vector per patch. A typical whole-slide image yields between 500 and 50{,}000 patches depending on tissue area, with each patch represented as a compact numerical vector rather than raw pixel data.

The current dataset contains pre-extracted features for approximately 32{,}000 cases comprising over 66{,}000 H\&E whole-slide images. Features are organized in a sharded storage architecture: 16 HDF5 shard files with a JSON routing index that maps each case to its shard and enumerates its constituent slides. This design avoids scanning all files when loading a cohort---the routing index provides instant lookup, and individual shard files are opened lazily on first access. Within each shard, features are organized hierarchically by case and slide, preserving the clinical relationship between slides from the same case.

\paragraph{Why pre-extracted features matter.} The pre-extraction approach is more than a usability convenience. First, every model trained on the platform consumes the same feature representation, so reported results are directly comparable across strategies and across users---there is no confound from differing tile sizes, foreground detection heuristics, or encoder choices. Second, it removes the dominant cost of computational-pathology research: feature extraction over the entire dataset would consume thousands of GPU-hours that a typical research group cannot afford, whereas a downstream MIL training run consumes orders of magnitude less compute. Third, the resulting training jobs are short enough (minutes to a few hours) that the platform can run many of them in parallel on the autoscaled cluster, enabling cheap multi-strategy comparisons, large hyperparameter sweeps, and rapid iteration that would be impractical if each run had to re-extract features.

\paragraph{Variable-length collation.} A fundamental challenge in MIL training is that each slide produces a different number of patches, so standard batching does not apply. The platform handles this by padding each batch to the length of the longest slide in that batch and constructing a boolean mask that distinguishes real patches from padding. All downstream aggregators use this mask to ignore padded positions---for example, masked mean pooling sums only real patch vectors, and the attention mechanism assigns zero weight to padding. This approach avoids truncating large slides or wasting computation on unnecessary padding across batches.

\subsection{MIL Classification Strategies}

All strategies share a common two-stage architecture: first, an \textit{aggregator} reduces the variable-length set of patch features into a single fixed-size slide-level representation; then, a \textit{classification head} (a multi-layer perceptron) maps that representation to class probabilities. The platform supports four strategies that span the complexity spectrum (Table~\ref{tab:strategies}).

\begin{table}[h]
\centering
\caption{Available classification strategies.}
\label{tab:strategies}
\begin{tabular}{@{}lp{6.5cm}l@{}}
\toprule
\textbf{Strategy} & \textbf{Description} & \textbf{Best For} \\
\midrule
Pooling MLP & Averages or takes the maximum across all patches, then classifies. Simple and robust. & Small cohorts, baselines \\
\addlinespace
Attention MIL & Learns which patches are most important for the classification decision~\cite{ilse2018attention}. & Default choice \\
\addlinespace
CLAM & Extends attention MIL with instance-level patch supervision~\cite{lu2021data}. & Larger cohorts \\
\addlinespace
LoRA MIL & Applies lightweight low-rank adaptation~\cite{hu2022lora} to classification layers. & Transfer learning \\
\bottomrule
\end{tabular}
\end{table}

\paragraph{Aggregator architectures.} The simplest aggregators---mean pooling, max pooling, and their concatenation (mean-max pooling)---require no learned parameters and are effective baselines. Mean pooling averages all patch vectors into a single 1024-dimensional slide vector; max pooling selects the element-wise maximum; mean-max concatenates both into a 2048-dimensional representation that captures both average tissue characteristics and the most extreme feature activations.

The attention-based aggregator (ABMIL)~\cite{ilse2018attention} is the default and most frequently used. It consists of a two-layer attention network: a linear projection from the 1024-dimensional feature space to a 128-dimensional attention space, followed by a Tanh nonlinearity, dropout, and a final linear layer that produces a single attention score per patch. These scores are normalized via softmax across all patches in the slide (with padding masked out), and the slide representation is computed as the attention-weighted sum of patch features. This allows the model to learn which tissue regions are diagnostically relevant and focus the classification on those regions.

\paragraph{Classification head.} All strategies feed the aggregated slide representation through a multi-layer perceptron (MLP) consisting of one or more fully connected layers with ReLU activations and dropout regularization between layers, followed by a final output layer that produces one logit per class. The default configuration uses a single hidden layer of 128 units with 50\% dropout, providing sufficient capacity for most binary and multi-class classification tasks while limiting overfitting on small cohorts.

\paragraph{Attention MIL} is the default strategy. It pairs the ABMIL aggregator with the standard MLP head, learning which patches drive the classification while remaining interpretable---the attention weights can be projected back onto the slide to visualize which regions the model considers most important.

\paragraph{CLAM (Clustering-constrained Attention MIL)} extends attention MIL with instance-level supervision~\cite{lu2021data}. In addition to the bag-level classification loss, CLAM identifies the top-$k$ highest-attention and bottom-$k$ lowest-attention patches (default $k = 8$) and applies a secondary loss: the highest-attention patches are trained to be positive instances (diagnostically relevant) and the lowest-attention patches are trained as negative instances (background or non-informative). Each class has its own instance classifier—a lightweight per-class linear layer—that provides this supervision. The instance loss is weighted (default 30\% of total loss) and added to the bag classification loss. This encourages sharper, more clinically interpretable attention patterns, particularly on larger cohorts where the attention network has enough data to learn meaningful instance-level distinctions.

\paragraph{LoRA MIL} applies Low-Rank Adaptation (LoRA)~\cite{hu2022lora} to the classification layers. Rather than training full weight matrices, LoRA decomposes each weight update into two small low-rank matrices (default rank 8, scaling factor 16), dramatically reducing the number of trainable parameters. This is particularly useful for transfer learning scenarios where a model trained on one tissue type is adapted to another. The LoRA adapters can optionally target the attention network layers in addition to the classification head, providing fine-grained control over which parts of the model are adapted.

\subsection{Handling Class Imbalance}

Medical imaging cohorts are frequently imbalanced---for example, a study of rare tumor subtypes may have 10$\times$ more cases of the common type. The platform addresses this automatically at two levels:

\begin{itemize}[leftmargin=*]
    \item \textbf{Weighted sampling}: when the imbalance ratio exceeds 1.5$\times$, the training loop automatically oversamples minority classes so that each class is seen approximately equally often during training.
    \item \textbf{Label smoothing}: the loss function is softened slightly (default: 10\%) to prevent the model from becoming overconfident, which improves calibration on small datasets.
\end{itemize}

\subsection{Training Robustness}

Several mechanisms improve training stability on the small, variable-sized datasets typical of pathology research:

\begin{itemize}[leftmargin=*]
    \item \textbf{Patch dropout}: randomly removes 10\% of patches from each slide during training, acting as data augmentation and preventing the model from memorizing specific patch configurations.
    \item \textbf{Cosine warmup scheduling}: the learning rate starts at 1\% of its target value and increases linearly over a warmup period (the first 10\% of total epochs, minimum 1 and maximum 5 epochs), then follows a cosine annealing schedule that smoothly decays the rate to near zero by the final epoch. The warmup phase is critical for attention-based models, where large early gradients can destabilize the attention weight initialization. The platform also supports pure cosine decay (without warmup), step decay (reducing the rate by a fixed factor every third of training), or a fixed learning rate.
    \item \textbf{Optimizer selection}: the default optimizer is AdamW, which combines Adam's adaptive per-parameter learning rates with decoupled weight decay regularization---the standard choice for transformer-based architectures and a good match for the cosine-warmup schedule. The platform also supports plain Adam and SGD as alternatives when a simpler optimizer is preferred. All three receive the same learning-rate and weight-decay values configured for the job.
    \item \textbf{Multi-criterion early stopping}: training can be stopped automatically when the model plateaus (no improvement for 15 epochs), but also when overfitting is detected---specifically, when the gap between training and validation performance exceeds a configurable threshold, indicating the model is memorizing rather than generalizing. A minimum epoch guard (default 10 epochs) prevents premature stopping during the initial learning phase, which is important because the cosine warmup alone occupies 3--5 epochs and metrics are unreliable before the model has trained with the full learning rate. The improvement threshold (default 0.02) filters noise from discrete metrics---on small validation sets, metrics like AUROC have few possible values, and tiny fluctuations would otherwise reset the patience counter.
    \item \textbf{Checkpoint management}: the training loop saves a checkpoint at every epoch where the monitored metric (default: validation AUROC) improves, recording the model weights, optimizer state, and epoch number. This ensures that if training is stopped---whether by early stopping, time limits, or user intervention---the best model seen during training is preserved. Checkpoints are written to a persistent output mount on cloud storage, so model artifacts survive after the GPU compute instance is released.
\end{itemize}

\section{Automated Hyperparameter Tuning}
\label{sec:hptuning}

Selecting the right hyperparameters---learning rate, model size, regularization strength---has an outsized impact on model performance but requires ML expertise. CellDX AI Autopilot automates this through an iterative pairwise search that supports two interchangeable sampling strategies: an exhaustive \emph{grid} search and a seeded \emph{random} search.

\subsection{Approach}

Rather than testing all possible combinations of all parameters simultaneously (which grows exponentially), the search is organized into stages. Each stage optimizes a pair of related parameters while holding all others fixed at their current best values. After each stage, the winning values are locked in and the search proceeds to the next pair.

The recommended configuration uses three stages:

\begin{enumerate}[leftmargin=*]
    \item \textbf{Learning rate and model capacity} --- the learning rate is searched over values spanning two orders of magnitude (e.g., 5e-5 to 5e-3), paired with the attention dimension (64 to 256 units) or hidden layer size. These interact strongly---larger models need lower learning rates---and have the biggest impact on performance.
    \item \textbf{Regularization strength} --- head dropout (10\% to 60\%) and aggregator dropout (5\% to 40\%) are tuned together. These control overfitting at two distinct points in the model: the aggregator dropout regularizes the attention scoring, while head dropout regularizes the classification layers.
    \item \textbf{Fine-grained smoothing} --- label smoothing (0\% to 20\%) and weight decay (1e-3 to 1e-1). These have smaller individual effects but compound with earlier choices.
\end{enumerate}

A typical three-stage search with 3--4 candidate values per parameter produces approximately 30 trials (e.g., $4 \times 3 + 4 \times 3 + 3 \times 3 = 33$), compared to nearly 1{,}000 for exhaustive search over the same parameter space---a reduction of over 30$\times$.

\paragraph{Search method: grid vs. random.} The platform exposes the sampling strategy as a configuration parameter (\texttt{hp\_tuning.method}). The default \emph{grid} method enumerates each stage's full Cartesian product---predictable cost, no missed combinations, well suited to small per-stage grids. When a stage has many candidate values (e.g., four parameters or four-plus values per parameter), the per-stage Cartesian product can grow uncomfortably large; the \emph{random} method addresses this by drawing \texttt{n\_trials\_per\_stage} configurations uniformly without replacement from the same Cartesian product, seeded by \texttt{hp\_tuning.seed} for reproducibility. Stage seeds are offset by the stage index so different stages explore different points even with the same root seed. When the requested random budget meets or exceeds a stage's grid size, the search falls back to the full grid rather than duplicating trials. The agent skill recommends grid for narrow stages ($\le 6$ combinations) and random when budget caps are needed; both methods plug into the same staged ``lock-the-winner'' pipeline and are recorded as part of each anonymous tuning outcome to inform future default updates.

\paragraph{Trial isolation and efficiency.} Each trial runs as an independent short training job with forced early stopping: a reduced patience (default 10 epochs) and a minimum epoch floor (default 5 epochs) that is lower than in full training. This aggressively terminates unpromising configurations after a few epochs while giving each trial enough time to reveal its learning trajectory. Between trials, GPU memory is explicitly released to prevent accumulation from affecting subsequent runs. The primary selection metric for all stages is validation AUROC, consistent with the final evaluation protocol.

The agent manages this entire process automatically: it constructs the stage definitions based on the chosen strategy, picks the search method appropriate to those stages, submits all trials, monitors their progress, selects the best configuration from each stage, and reports the final optimized hyperparameters to the user.

\paragraph{Anonymous outcome telemetry.} On successful HP-tuning runs, the platform records an anonymized record of the outcome: the strategy, search method, winning hyperparameter values (filtered through a strict allowlist that excludes any cohort, label, or user identifiers), the configured baseline values, the winning metric value, and a one-way SHA-256 hash of the job identifier for deduplication. These records are stored in a separate database collection and exposed through a read-only endpoint, providing the raw material for periodic offline benchmark studies that can update the platform's default hyperparameters without requiring access to any user data.

\subsection{Multi-Strategy Comparison}

After hyperparameter tuning, the platform supports automated comparison of all four strategies on the same cohort and hyperparameters. The agent submits parallel training jobs, monitors them, and presents a comparison table when all complete. This enables the user to make an informed deployment decision without understanding the technical differences between strategies.

\section{Evaluation and Metrics}
\label{sec:evaluation}

The platform tracks a comprehensive set of metrics at every training epoch:

\begin{itemize}[leftmargin=*]
    \item \textbf{AUROC} (Area Under the ROC Curve) --- the primary model selection metric, measuring discrimination ability across all classification thresholds.
    \item \textbf{PR-AUC} (Precision-Recall AUC) --- particularly informative for imbalanced datasets where high AUROC can coexist with poor minority-class detection.
    \item \textbf{Balanced accuracy} --- average per-class recall, providing an unbiased estimate when class sizes differ significantly.
    \item \textbf{Macro F1, precision, and accuracy} --- additional metrics reported for completeness.
\end{itemize}

All metrics use macro-averaging, giving equal weight to each class regardless of its frequency. This is essential in pathology where disease subtypes of clinical interest may be rare in the cohort. Metrics are computed from softmax probabilities at every epoch for both training and validation sets, providing a continuous view of model behavior rather than only final-epoch results.

For rigorous evaluation, the platform supports stratified $k$-fold cross-validation (default: $k = 5$), which preserves class proportions in each fold and reports mean $\pm$ standard deviation across folds. Each fold trains a complete model from scratch with the same hyperparameters, ensuring that reported performance reflects the stability of the configuration rather than a lucky split. All experiments use deterministic seeding for full reproducibility: the same seed, cohort, and configuration will produce identical results across runs.

\paragraph{Held-out test set policy.} The agent skill enforces a stratified \mbox{75 / 15 / 15} train / validation / test split as the default for every cohort, with a non-empty test set required whenever a 15\% test fraction yields at least five slides per class. The validation set is used for HP tuning, early stopping, and best-checkpoint selection; the test set is held out and reported once on the final selected checkpoint. This separation is necessary because the validation metric is implicitly overfit through repeated model selection, and only the held-out test set provides an unbiased estimate of generalization. Smaller cohorts that cannot support a meaningful test set fall back to an 80 / 20 train / validation split, with the agent explicitly informing the user that no held-out generalization estimate was produced.

\section{Deployment}
\label{sec:deployment}

The platform enforces a strict separation between training and deployment. After a model is trained and the user reviews its performance, deployment requires an explicit API call with user-facing metadata: a title, description, and target organ. The agent guides this process but cannot deploy without the user's approval.

Deployed models are packaged with two artifact files: a \textit{model configuration} (specifying the class labels, aggregator type, attention dimension, hidden layer sizes, dropout rates, and feature dimensionality---everything needed to reconstruct the model architecture) and the \textit{trained weights} checkpoint (containing the model state dictionary, optimizer state, and best epoch metadata). Together, these artifacts allow exact reconstruction and loading of the trained model without access to the original training configuration or code. The artifacts are persisted to blob storage at a deterministic path derived from the job identifier, ensuring they remain available after the GPU compute instance is released.

At inference time, new slides are processed through the same pipeline: tissue detection, non-overlapping patch extraction at 224$\times$224 pixels, feature extraction via the same Hibou-L encoder that produced the training features, variable-length bag collation, and MIL classification using the reconstructed model. The inference module reconstructs the exact aggregator and classification head from the model configuration file, loads the trained weights, and produces per-class probabilities along with the predicted class label.

The deployment response includes only clinical metadata (model name, performance summary, organ, tags)---no internal architecture details are exposed. This prevents information leakage and keeps the user interface focused on what pathologists care about: what the model classifies, how well it performs, and for which tissue type.

\section{Limitations and Future Work}
\label{sec:limitations}

We acknowledge several important limitations of the current system, and outline the concrete directions we plan to pursue.

\paragraph{H\&E classification only.} The platform currently supports only H\&E-stained whole-slide images for classification tasks. It does not yet support immunohistochemistry (IHC) stains, special stains, fluorescence imaging, or non-classification tasks such as segmentation, detection, or regression.

\paragraph{Fixed feature extractor.} The system uses pre-extracted features from a single foundation model (Hibou-L). While this dramatically simplifies the user experience, it means that the quality of downstream classifiers is bounded by the feature extractor's capabilities. The platform does not currently support end-to-end fine-tuning of the feature extractor or integration of alternative foundation models (UNI~\cite{chen2024uni}, Virchow~\cite{vorontsov2024virchow}, Prov-GigaPath~\cite{xu2024gigapath}).

\paragraph{No spatial modeling.} The MIL strategies treat each slide as an unordered bag of patches. They do not model spatial relationships between patches, which may be important for tasks where tissue architecture is diagnostically relevant (e.g., gland formation patterns, tumor invasion fronts). Graph-based and hierarchical approaches~\cite{chen2022hipt} are not yet supported.

\paragraph{Pairwise tuning assumptions.} The iterative pairwise hyperparameter search assumes limited interaction between parameter groups across stages. While this holds well in practice for the current parameter set, it could miss complex three-way interactions. Bayesian optimization or reinforcement learning-based approaches may be explored in future versions, and the anonymous outcome telemetry described in Section~\ref{sec:hptuning} provides the empirical basis for periodically re-tuning the platform's default values.

\paragraph{Agent skill limitations.} The current agent skills are structured documents that guide LLM behavior but do not provide formal guarantees. The agent's ability to correctly interpret training results and make appropriate recommendations depends on the underlying LLM's capabilities. Systematic evaluation of agent-guided training outcomes compared to expert-configured baselines is ongoing work.

\paragraph{Dataset scope.} While the platform provides access to over 32{,}000 cases and 66{,}000 slides, these are currently limited to H\&E-stained tissue. The feature store does not yet cover all tissue types or pathology subspecialties uniformly.

\subsection{Planned Directions}
\label{sec:future-work}

The following directions are actively planned for the next development cycles.

\paragraph{Scaling the pre-processed feature store toward one million slides.} The sharded HDF5 layout described in Section~\ref{sec:training} was designed with horizontal scaling in mind: adding capacity is a matter of producing additional shard files and extending the JSON routing index. We plan to grow the feature store toward one million pre-processed whole-slide images, broadening organ and indication coverage so that the agent can validate sufficient cohort sizes for a much wider range of classification objectives without users needing to upload or process slides themselves.

\paragraph{IHC stain support and the labeling cost it implies.} Adding immunohistochemistry (IHC) stains is a natural extension of the platform---once IHC features can be extracted, the same MIL training pipeline applies. The non-trivial cost is labeling: IHC slides routinely include control tissues (e.g., positive and negative tonsil cores) on the same physical slide as the patient sample, and these controls must be annotated and excluded before training, otherwise the classifier will learn to discriminate based on the controls rather than the diagnostic tissue. This requires region-level annotation by a pathologist for every slide ingested, which is a substantial manual effort that does not exist for H\&E. We plan to introduce a region-of-interest annotation workflow specifically to support reliable IHC training, gated by the cost-benefit trade-off for each indication.

\paragraph{Segmentation through publicly labeled datasets.} The platform currently supports only slide-level classification. Public datasets that pair pathology images with dense pixel-level annotations---most notably HistAI's SPIDER dataset---make it feasible to add semantic segmentation as a first-class task type alongside classification. Concretely, we plan to extend the agent skill set with a segmentation skill that consumes the same patch-feature representations but trains a per-patch (or per-pixel, after upsampling) classifier head, enabling tasks such as tumor delineation, mitotic figure detection, and gland boundary extraction. Reusing already-labeled public corpora bootstraps the segmentation capability without requiring users to provide their own pixel-level labels.

\paragraph{Custom datasets and validation on independent public datasets.} Today the platform exposes a single curated feature store. We plan to allow users to register their own datasets---uploading or referencing slides, running feature extraction in an isolated batch step, and reusing the resulting shards in the same training pipeline---so that institutional cohorts with non-public tissue can be trained on without leaving the platform. In parallel, we plan to add automated validation against established public benchmarks (e.g., TCGA cohorts, CAMELYON, PANDA) so that any model trained through the platform can be reported with both an internal held-out test metric and an independent external metric, strengthening claims of generalization beyond the user's own institution.

\section{Discussion}

The central thesis of CellDX AI Autopilot is that the bottleneck for AI adoption in pathology is not algorithmic---the methods exist---but operational. Most pathologists who could benefit from custom classifiers do not have the ML skills to build them, and most ML teams do not have the pathology expertise to know which classifiers would be clinically valuable. By placing an AI agent between these two worlds, the platform aims to unlock a much larger number of pathology AI applications than is possible with the current expert-dependent workflow.

This approach differs fundamentally from both commercial pathology AI (which offers fixed, pre-built models) and general-purpose AutoML (which lacks domain knowledge). The agent skill architecture allows encoding pathology-specific best practices---appropriate metrics, minimum cohort sizes, clinical terminology requirements, mandatory human approval for deployment---directly into the system's operational logic.

The choice to operate on pre-extracted features, while limiting, is deliberate. It eliminates the single largest barrier to entry: the need for GPU infrastructure and expertise to run feature extraction on gigapixel images. A pathologist with a classification question and a curated cohort can go from hypothesis to trained model in hours rather than weeks.

We see CellDX AI Autopilot as a step toward a broader vision: AI-assisted AI development for specialized scientific domains, where domain experts drive the research questions and AI agents handle the engineering complexity.

\section{Conclusion}

We have presented CellDX AI Autopilot, a platform that lets users---from pathologists with no ML background to ML practitioners running many parallel experiments---train, evaluate, and deploy whole-slide image classifiers through AI-agent-guided interaction. The platform combines structured agent skills with automated MIL training infrastructure, supporting four classification strategies, automated hyperparameter tuning with a 30$\times$ reduction in search cost (grid or seeded random), and human-in-the-loop deployment. Operating on a pre-built feature store of over 32{,}000 cases and 66{,}000 H\&E slides offered free to platform users, the system addresses two coupled bottlenecks: the ML-expertise barrier that limits adoption in diagnostic pathology, and the engineering / data-cost barrier that limits how many experiments a researcher can run affordably. To our knowledge, this is the first system to expose a pathology-specialized agent-skill set and a pathology-specialized training platform to general-purpose AI agents, delivering end-to-end automated pathology model training and deployment without requiring the agent runtime itself to be domain-specific.

\section*{Disclosure}

This manuscript was drafted with the assistance of Claude Code (Anthropic) acting as an authoring tool. All technical content, claims, and numerical figures were reviewed, edited, and approved by the authors prior to submission.

\bibliographystyle{plain}

\end{document}